%% file: main.tex
\documentclass[10pt,twocolumn,letterpaper]{article}

\usepackage[pagenumbers]{cvpr} 

\input{preamble}

\definecolor{cvprblue}{rgb}{0.21,0.49,0.74}
\usepackage[pagebackref,breaklinks,colorlinks,citecolor=cvprblue]{hyperref}

\usepackage{multirow}
\usepackage{bbding}
\usepackage{color}
\usepackage{enumerate}

\def\paperID{5503} 
\def\confName{CVPR}
\def\confYear{2024}

\title{DeepFidelity: Perceptual Forgery Fidelity Assessment for Deepfake Detection}

\author{Chunlei Peng\textsuperscript{1}, Huiqing Guo\textsuperscript{1}, Decheng Liu\textsuperscript{1}, Nannan Wang\textsuperscript{1}, Ruimin Hu\textsuperscript{1}, Xinbo~Gao\textsuperscript{2}\\
\textsuperscript{1}Xidian University, \textsuperscript{2}Chongqing University of Posts and Telecommunications\\
}

\begin{document}
\maketitle
\input{sec/0_abstract}    
\input{sec/1_intro}
\input{sec/2_related_work}
\input{sec/3_proposed_method}
\input{sec/4_experiments}
\input{sec/5_conclusion}
{
    \small
    \bibliographystyle{ieeenat_fullname}

}


\end{document}

%% file: preamble.tex
\usepackage[dvipsnames]{xcolor}


%% file: sec/0_abstract.tex
\begin{abstract}
Deepfake detection refers to detecting artificially generated or edited faces in images or videos, which plays an essential role in visual information security. Despite promising progress in recent years, Deepfake detection remains a challenging problem due to the complexity and variability of face forgery techniques. Existing Deepfake detection methods are often devoted to extracting features by designing sophisticated networks but ignore the influence of perceptual quality of faces. Considering the complexity of the quality distribution of both real and fake faces, we propose a novel Deepfake detection framework named DeepFidelity to adaptively distinguish real and fake faces with varying image quality by mining the perceptual forgery fidelity of face images. Specifically, we improve the model's ability to identify complex samples by mapping real and fake face data of different qualities to different scores to distinguish them in a more detailed way. In addition, we propose a network structure called Symmetric Spatial Attention Augmentation based vision Transformer (SSAAFormer), which uses the symmetry of face images to promote the network to model the geographic long-distance relationship at the shallow level and augment local features. Extensive experiments on multiple benchmark datasets demonstrate the superiority of the proposed method over state-of-the-art methods. The code is available at \url{https://github.com/shimmer-ghq/DeepFidelity}.
\end{abstract}

%% file: sec/1_intro.tex
\section{Introduction}
\label{sec:intro}

\begin{figure}
  \centering
   \includegraphics[width=1.0\linewidth]{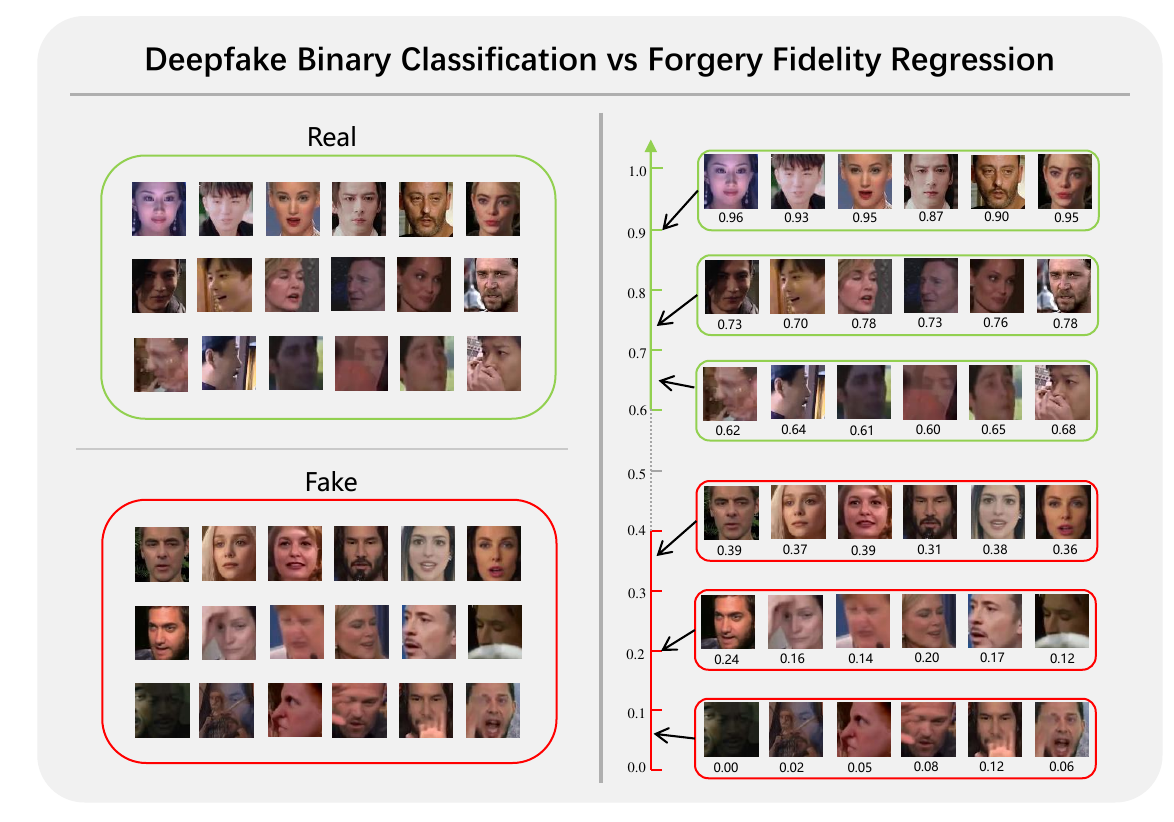}
   \caption{Deepfake binary classification vs forgery fidelity regression. Left subfigure shows the traditional Deepfake detection strategy based on binary real/fake classification, and right subfigure illustrates our perceptual forgery fidelity assessment approach, which could not only classify the real and fake faces, but also provide the forgery fidelity of them.}
   \label{fig:classification-regression}
\end{figure}

The rapid development of face forgery technology has brought about increasingly serious security risks. Malicious use of forged faces can lead to severe problems such as false authentication, information leakage, and financial fraud and even have far-reaching effects on society and individuals. In this context, forged face detection is critical. It is a crucial line of defense to protect digital identity security and effectively prevents potential security risks by identifying and intercepting the use of forged faces. Therefore, research and development of reliable forged face detection techniques are crucial for maintaining the security and stability of the digital society.

Existing methods for fake face detection can be broadly classified into three categories: spatial clue-based methods, temporal clue-based methods, and generalizable methods. Spatial clue-based methods~\cite{zhu2021face, chen2021local, shiohara2022detecting, wang2023dynamic} usually rely on image/frame level true/false classification to tap into the face's inherent features or the image's forged traces. Temporal clue-based methods~\cite{hosler2021deepfakes, hu2022finfer, haliassos2022leveraging, wang2023altfreezing} utilize consecutive video frames to mine spatiotemporal inconsistencies in fake videos, such as unnatural facial movements and temporal artifacts. Generalizable methods~\cite{fung2021deepfakeucl, sun2021improving, cao2022end, dong2023implicit} attempt to improve the model's generalization ability by extracting more robust representations.

Considering the discrepancy in creating procedures between real and fake faces, assessing the quality of the faces could be an easy option for forged face detection. Image quality assessment (IQA) based methods~\cite{galbally2014face, li2018can, korshunov2018deepfakes, tian2022generalized} have been applied to face forgery detection and face anti-spoofing tasks, which attempt to explore the quality differences between real and fake faces by performing quality assessments. However, with the development of Deepfake generation techniques, the difference in image quality distribution between real face images and high-quality fake face images becomes increasingly difficult to distinguish. Therefore, methods based on IQA could not perform well on Deepfake detection. This inspired us to recognize that while the quality distribution between real and forged face data may not show significant differences, substantial quality variations exist within each set of real and forged face data. Low-quality face images, including blurriness, occlusions, and challenging angles, tend to lead to a partial loss of forgery clues. Therefore, in forged face detection tasks, the presence of face data of varying qualities may result in differences in detection difficulty, posing a challenge for model training. In view of this, we introduce the concept of perceptual forgery fidelity, in which the former two-classes score is replaced by the proposed forgery fidelity score ranging from 0 to 1. Different from IQA, the fidelity score is based on the authenticity of face data while also considering the quality of the image or face. How to explore and assess the perceptual forgery fidelity between real and fake faces has become an important and interesting question, which we will answer below.

In this paper, we present a novel Deepfake detection method, namely DeepFidelity, based on assessing the perceptual forgery fidelity of face images, which provides a simple and robust solution for Deepfake detection.

Firstly, for perceptual feature extraction, we design a Symmetric Spatial Attention Augmentation based vision Transformer called SSAAFormer. Specifically, we employ convolution operations at the shallow level of the network to capture low-level features of the image. Considering the symmetry of face images, we introduce the geographic long-distance relationship convolution to replace conventional convolutions, enhancing local features by leveraging the symmetry of face data. At the deep level of the network, we employ the self-attention mechanism to model global dependencies. This not only helps improve computational efficiency but also aids the model in learning more expressive feature representations.

Finally, we employ a novel strategy to map real and fake face data of different qualities to distinct scores. This enables us to distinguish face data of varying qualities more comprehensively during detection, enhancing the model's ability to recognize complex samples. This strategy provides an effective means to improve the performance of our deepfake detection model, particularly in handling face data of diverse qualities.

The main contributions of our work are summarized as follows:
\begin{itemize}
\item[$\bullet$]
Starting from a novel perspective, we propose the DeepFidelity framework for face forgery detection, which improves the model's capability to recognize complex samples by mapping face data of different qualities to distinct scores, allowing for a more comprehensive and fairer differentiation.
\item[$\bullet$]
We propose the Symmetric Spatial Attention Augmentation based vision Transformer (SSAAFormer), which leverages the geographic long-distance relationship of face data to facilitate more effective learning of facial representations at the shallow level of the model.
\item[$\bullet$]
Experiments on benchmark datasets, including FaceForensics++, Celeb-DF (v2), and WildDeepfake, as well as cross-dataset evaluations, help validate the superiority and generalizability of the proposed method compared with state-of-the-art methods.
\end{itemize}

%% file: sec/2_related_work.tex
\section{Related Work}
\label{sec:related_work}

In this section, we briefly review representative Deepfake detection methods in the following categories: spatial clue-based methods, temporal clue-based methods, generalizable methods, and IQA-based methods.

\noindent
\textbf{Spatial Clue-based Deepfake Detection Methods.}
Zhang \etal~\cite{zhang2017automated} used SVM, random forest, and MLP for static face-swapping image detection, which is an early approach to solving the problem of face-swapping image detection using classical machine learning methods. Zhou \etal~\cite{zhou2017two} proposed a two-stream network for face tampering detection, where one stream detects low-level inconsistencies between image patches and another stream explicitly detects tamper faces. He \etal~\cite{he2019detection} employed residual signals of chrominance components from multi-color spaces (YCbCr, HSV, and Lab) as the input of CNN to extract common features. Zhu \etal~\cite{zhu2021face} mined forgery clues from the 3D decomposition of face images to construct facial details to amplify subtle artifacts. Chen \etal~\cite{chen2021local} fused RGB and frequency information to learn similarity patterns between local regions to distinguish real regions from forged regions and capture forged traces. Ni \etal~\cite{ni2022core} used cosine similarity loss as consistency loss to explicitly constrain the consistency representation of different augmented images. Mazaheri \etal~\cite{mazaheri2022detection} performed facial expression manipulation detection using a facial expression recognition system to locate manipulated regions in forged faces. Wang \etal~\cite{wang2023dynamic} utilizes frequency-aware information to assist CNNs in discovering forgery clues, using a spatial-frequency dynamic graph model to mine subtle forgery clues from relationships in the spatial and frequency domains.

\noindent
\textbf{Temporal Clue-based Deepfake Detection Methods.}
Considering that image noise is susceptible to corruption in compressed videos, Afchar \etal~\cite{afchar2018mesonet} used deep neural networks with a low number of layers to focus on the mesoscopic properties of images. Güera \etal~\cite{guera2018deepfake} used a CNN to extract frame-level features and used a simple convolutional LSTM to learn to classify if a video has been manipulated or not. \cite{sabir2019recurrent} was proposed after \cite{guera2018deepfake}, and unlike it, Sabir \etal used facial alignment to remove confounding factors when detecting facial manipulations and made use of bidirectional recurrency instead of mono-directional. Montserrat \etal~\cite{montserrat2020deepfakes} proposed an automatic weighting mechanism to emphasize the most reliable regions where faces have been detected and discard the least reliable ones when determining a video-level prediction. Hosler \etal~\cite{hosler2021deepfakes} exploited the inconsistency of the emotions conveyed by video and audio to detect deep forgery videos. Hu \etal~\cite{hu2022finfer} proposed a deep forgery detection method based on frame inference. The facial representations of future frames are predicted using facial representations of current frames and the deep forgery video is detected based on the correlation of facial representations between predicted future frames and referenced future frames. To capture spatial and temporal artifacts, Wang \etal~\cite{wang2023altfreezing} proposed a new training strategy that alternately trains spatial and temporal weights to enable the model to learn spatio-temporal features.

\noindent
\textbf{Generalizable Deepfake Detection Methods.} Cozzolino \etal~\cite{cozzolino2018forensictransfer} introduced a new representation learning method to facilitate transferability between image manipulation domains based on auto-encoders. Du \etal~\cite{du2020towards} proposed a locality-aware auto-encoder to bridge the generalization gap, using extra pixel-wise forgery masks for regularization to learn meaningful and intrinsic forgery representations. Considering the intrinsic image discrepancies across the blending boundary of fake face images, Li \etal~\cite{li2020face} proposed a novel representation called face X-ray, which was capable of distinguishing unseen forged images and predicting the blending regions. Luo \etal~\cite{luo2021generalizing} proposed to exploit high-frequency noise of images to improve the generalization ability of forgery detection. Sun \etal~\cite{sun2021improving} improved the effectiveness and robustness of forgery detection by integrating facial landmarks and their temporal dynamic features. Cao \etal~\cite{cao2021metric} used paired images with different compression levels and combined adversarial learning and metric learning to extract compression-insensitive features. The feature distance between compressed and uncompressed forgery images was reduced to improve the ability to handle compressed face forgery. Cao \etal~\cite{cao2022end} captured essential differences between real and fake faces by reconstructing real faces and mining the generic representation of the faces. Dong \etal~\cite{dong2023implicit} found that the sensitivity of the deepfake detection model to the data's identity information reduces the model's generalization ability. For this reason, they proposed an ID-unaware deepfake detection model to alleviate the Implicit Identity Leakage phenomenon.

\noindent
\textbf{IQA-based Forgery Detection Methods.} IQA is an easy solution that has been applied in the fields of face forgery detection and face anti-spoofing. A face anti-spoofing method based on IQA was proposed in \cite{galbally2014face}, which took a gray-scale image and applies Gaussian filtering to it. The original gray-scale image with the filtered image was then fed into a full-reference IQA model to perform detection. The principle of this method was to assume that the quality loss produced by the real image and the fake image after Gaussian filtering would be different. Li \etal~\cite{li2018can} used the face quality assessment method in \cite{chen2014face} to obtain quality scores of real and forged faces. However, since the face quality assessment only considers the appearance of the face and the difference between forged and real face images is not significant, the obtained scores have only a small difference in the distribution between the real and forged data. Korshunov \etal~\cite{korshunov2018deepfakes} proposed to use image quality measurement and SVM for the detection of forged faces, which only focused on the imaging modality, as in both \cite{galbally2014face} and \cite{li2018can}. Tian \etal~\cite{tian2022generalized} proposed to use the IQA method for objective evaluation to generate pseudo-labels and human opinion scores as a subjective evaluation method. This method tended to focus on human visual perception, which cannot be applied to discriminate between real and forged images when the forged images are too realistic to be distinguished with the naked eye.

As mentioned above, existing Deepfake detection methods either capture spatial or temporal artifacts in forged face videos or employ sophisticated networks to extract robust features. Most IQA-related detection methods focus on human visual perception, but the quality distribution difference between real and fake face images is not significantly distinct. Different from existing studies, our proposed DeepFidelity method aims to more comprehensively and fairly distinguish real and fake face data with different face qualities by mapping them to distinct scores. Furthermore, we leverage the geographic long-distance relationship of face data to facilitate effective facial representation learning at the shallow level of the network.

%% file: sec/3_proposed_method.tex
\section{Proposed Method}
\label{sec:proposed_method}

\begin{figure*}
  \centering
    \includegraphics[width=0.9\linewidth]{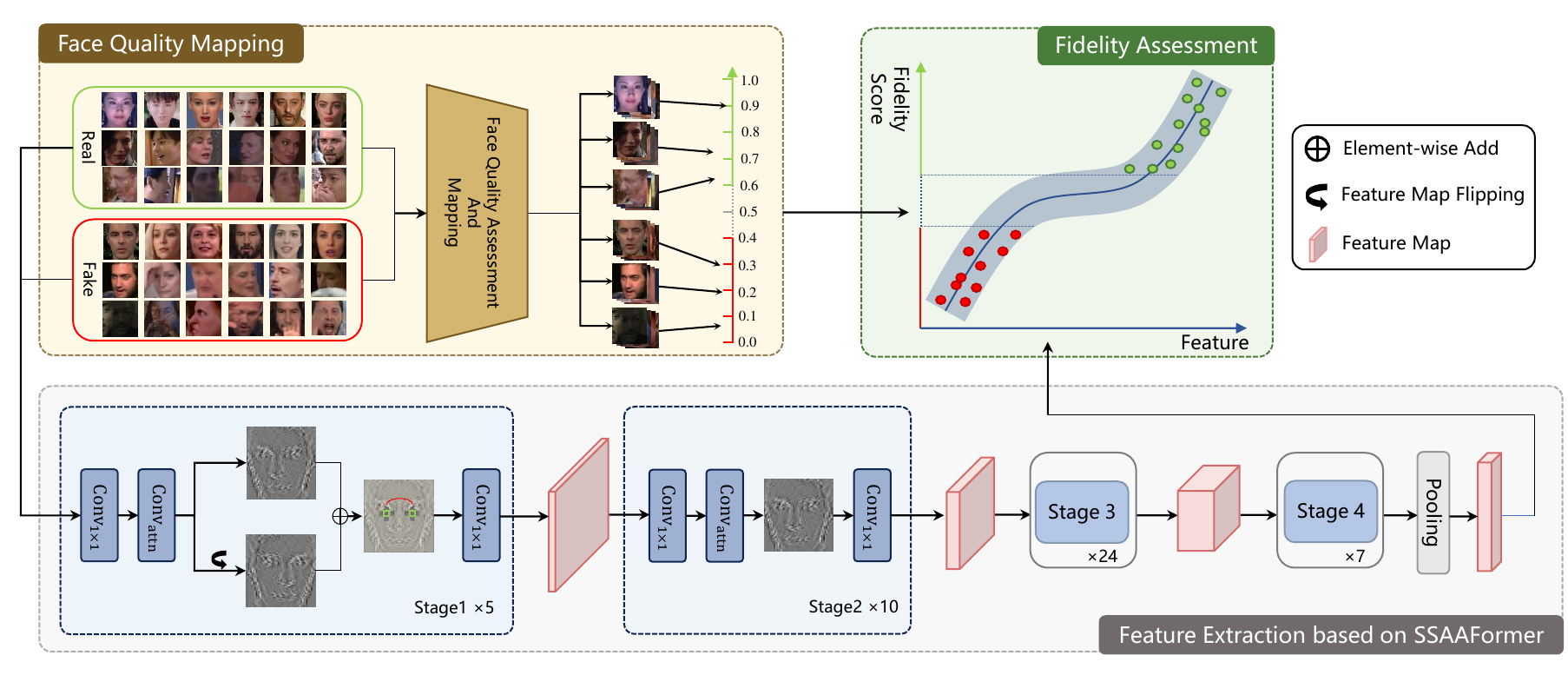}
  \caption{Overview of the proposed perceptual forgery fidelity assessment for Deepfake detection. The face quality mapping phase could distinguish real and fake faces more precisely by mapping them to varying fidelity scores based on their image quality. The feature extraction network is designed with proposed symmetric spatial attention augmentation to assess the perceptual fidelity of the face images.}
  \label{fig:framework}
\end{figure*}

In this section, we present details of the proposed DeepFidelity approach. Considering the complexity of face data quality distribution in the real world, we introduce the concept of perceptual forgery fidelity. Specifically, by mapping face data of different qualities to distinct scores, we replace the previous binary classification with the forgery fidelity score ranging from 0 to 1. Then we design a network architecture called SSAAFormer to assess the perceptual forgery fidelity of input images, which augments the local features extracted at the shallow level of the network by exploiting the inherent symmetry in face data to model the geographic long-distance relationship. \Cref{fig:framework} gives an overview of the proposed method.

\subsection{Perceptual Forgery Fidelity Assessment}

One key observation is that low-quality face images, including blurriness, occlusions, and challenging angles, often lead to a partial loss of forgery clues, making it difficult for deepfake detection models to distinguish their authenticity. The varying quality of face data can indeed pose a challenge to the model training. Based on this observation, we aim to distinguish different quality face data in more detail by mapping them into distinct scores.

Specifically, we first use the previously published method for face quality assessment~\cite{terhorst2020ser} to score each face image. Assuming we simply map face images of different qualities to different scores, there would be no distinction between real and fake faces. Therefore, we normalize the quality scores of fake and real face images into different ranges. To make the scores of real and fake data more distinguishable, we set the lower limit for real face data to 0.6 and the upper limit for fake face data to 0.4. As shown in \cref{fig:classification-regression}, the green portion represents real face data, the red portion represents fake face data, and face data of different qualities are mapped into different score ranges.

\subsection{SSAAFormer}

\begin{figure}
  \centering
   \includegraphics[width=1.0\linewidth]{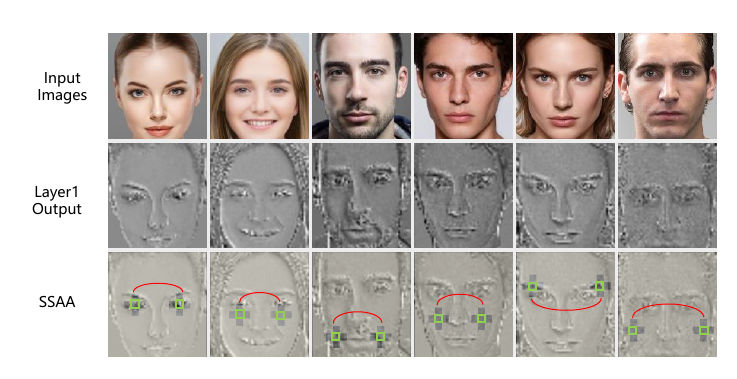}
   \caption{Illustration of the proposed symmetric spatial attention augmentation (SSAA) performed on the feature maps in the shallow layers.}
   \label{fig:ssaa}
\end{figure}

Following the popular ViT architecture, SSAAFormer consists of four stages, each containing $ N_i $ stacked blocks.

It is shown that for a given token, the spatial attention of ViTs at the shallow level primarily focuses on its neighboring tokens~\cite{li2023uniformer}. However, this is obtained by comparing all tokens, which undoubtedly leads to a significant amount of redundant computation. Therefore, following \cite{li2023uniformer}, we use convolution operations in the first two stages of the network instead of self-attention mechanism to capture local features and positional information. Specifically, the input feature map $ X $ is first subjected to non-linear transformations among channel dimensions using a $ 1\times1 $ convolution. Then, convolution operations are performed on each channel separately to learn the contextual relationships between each token and its surrounding adjacent tokens.

\begin{equation}
  A\left(X\right)={{Conv}}{}_{{attn}}\left({{Conv}}{}_{{1\times1}}\left(X\right)\right),
  \label{eq:1}
\end{equation}

In convolutional operations, the size of the receptive field is fixed. This means that convolutional layers can only capture local information and cannot access global contextual information. Therefore, as shown in \cref{fig:ssaa}, in the first stage of the network, we leverage the symmetry of face data to model the geographic long-distance relationship, augmenting local features to generate more representative feature representations. Specifically, we perform a flipping operation on $ A(X) $ to obtain its mirrored feature, $A_f(X)$. Then we set two learnable parameters, $ w1 $ and $ w2 $, to weight and sum of $ A(X) $ and $A_f(X)$ to obtain the symmetric spatial attention augmented feature, $ SSAA(X) $. Next, feature correlations are performed in the channel dimensions using a $ 1\times1 $ convolution.

\begin{equation}
  {{A}}{}_{{f}}\left(X\right)\xleftarrow{Flip}A\left(X\right),
  \label{eq:2}
\end{equation}

\begin{equation}
  SSAA\left(X\right)={{w}}{}_{{1}}\cdot A\left(X\right)\oplus{{w}}{}_{{2}}\cdot{{A}}{}_{{f}}\left(X\right),
  \label{eq:3}
\end{equation}

\begin{equation}
  C\left(X\right)={{Conv}}{}_{{1\times1}}\left(SSAA\left(X\right)\right)
  \label{eq:4}
\end{equation}

As the number of convolutional layers in the network increases, the intermediate features no longer exhibit significant geographic long-distance relationships. Therefore, we did not employ symmetric feature augmentation in stage2.

\begin{equation}
  C\left(X\right)={{Conv}}{}_{{1\times1}}\left(A\left(X\right)\right)
  \label{eq:5}
\end{equation}

To enable the model to perform contextual analysis on features at both local and global scales, we introduce self-attention modules in the third and forth stages of the network. Inspired by \cite{li2021localvit}, we employ a deep convolutional layer for dynamic position embedding, enabling the model to better understand the importance of different positions in the input sequence. Then the input $ X $ undergoes three different linear transformations to obtain queries (Q), keys (K), and values (V), which are then fed into a multi-head attention module for feature aggregation among tokens. Subsequently, the aggregated features pass through a feed-forward network consisting of two fully connected layers.

\begin{equation}
  Q,K,V\xleftarrow{Linear}X,
  \label{eq:6}
\end{equation}

\begin{equation}
  Attention\left(Q,K,V\right)=soft\max\left(\frac{QK^T}{\sqrt{{{d}}{}_{{k}}}}\right)V
  \label{eq:7}
\end{equation}


Following this, we employ Support Vector Regression (SVR) to map features to scores. We utilize the Radial Basis Function (RBF) as the kernel:

\begin{equation}
    K\left(x,x^{^{\prime}}\right)=\exp\left(-\frac{\Vert{x-x^{^{\prime}}}\Vert^2}{2\sigma^2}\right),
  \label{eq:8}
\end{equation}

\noindent
where $ \sigma $ is a hyperparameter of the RBF kernel, controlling the rate of decay of similarity between samples.

%% file: sec/4_experiments.tex
\section{Experiments}
\label{sec:experiments}


\begin{table*}
\begin{center}
\renewcommand{\arraystretch}{1.2}
\belowrulesep=0pt
\aboverulesep=0pt
\begin{tabular}{cccccccccc}
\toprule
\multirow{2.5}{*}{Method} &
\multicolumn{2}{c}{FF++ (HQ)} &
 &
\multicolumn{2}{c}{WildDeepfake} &
 &
\multicolumn{2}{c}{Celeb-DF (v2)} \\
\cline{2-3}
\cline{5-6}
\cline{8-9}
  & Acc ($\%$) & AUC ($\%$) &  & Acc ($\%$) & AUC ($\%$) &  & Acc ($\%$) & AUC ($\%$) \\
\midrule
Xception \cite{chollet2017xception} & 95.73 & 96.30 &  & 79.99 & 88.86 & & 97.90 & 99.73 \\
EfficientNet-b4 \cite{tan2019efficientnet} & 96.63 & 99.18 & & 82.33 & 90.12	& & 98.19 & 99.83 \\
Add-Net \cite{zi2020wilddeepfake} & 96.78 &	97.74 & & 76.25 & 86.17 & &	96.93 &	99.55  \\
F3Net \cite{qian2020thinking} & 97.52 & 98.10 & &	80.66 &	87.53 & & 95.95 & 98.93  \\
SCL \cite{li2021frequency} &	96.69 &	99.30 & & - & - & &	- & -  \\
MADD \cite{zhao2021multi} & 97.60 & 99.29 & & 82.62 & 90.71 & & 97.92 & 99.94  \\
Local Relation \cite{chen2021local} & 97.59 & 99.46 & & - & - & &	- &	-  \\
PEL \cite{gu2022exploiting}	& 97.63 & 99.32 & & 84.14	& 91.62 & &	- &	-  \\
RECCE \cite{cao2022end} & 97.06 & 99.32 & &	83.25 &	92.02 & & 98.59 & 99.94  \\
M2TR \cite{wang2022m2tr} & \underline{98.23} & 99.48 & & - & - & & - & -  \\
SFGD (Xception) \cite{wang2023dynamic} &	97.61 &	99.45 & & 83.36 & 92.15 & &	98.95 &	99.94  \\
SFGD \cite{wang2023dynamic} & 98.19 & \underline{99.53} & & \underline{84.41} & \textbf{92.57} & & \underline{99.22} & \underline{99.96}  \\
\midrule
\textbf{Ours} & \textbf{98.89} & \textbf{99.67} & & \textbf{88.34} & \underline{92.33} &  & \textbf{100} & \textbf{100} \\
\bottomrule
\end{tabular}
\end{center}
\caption{Intra-dataset comparison in terms of Acc ($\%$) and AUC ($\%$). Comparison with state-of-the-art methods on FF++ (HQ), WildDeepfake, and Celeb-DF (v2). Best results are marked in bold, and the second best are marked with underline.}
\label{tab:intro}
\end{table*}

\begin{table}
  \centering
  \belowrulesep=0pt
  \aboverulesep=0pt
  \begin{tabular}{c|c|c}
    \toprule
    \multicolumn{1}{c|}{Training DataSet} & \multicolumn{2}{c}{FF++ (LQ)} \\
    \midrule
        Method & WildDeepfake & Celeb-DF (v2) \\
        \midrule
         Xception \cite{chollet2017xception} & 60.59 & 60.05 \\
         EfficientNet-b4 \cite{tan2019efficientnet} & 64.27 & 64.29 \\
         Add-Net \cite{zi2020wilddeepfake} & 54.21 & 57.83 \\
         F3Net \cite{qian2020thinking} & 60.49 & 67.95 \\
         RFM \cite{wang2021representative} & 57.75 & 65.63 \\
         LTW \cite{sun2021domain} & 67.12 & 64.10 \\
         Multi-att \cite{zhao2021multi} & 65.65 & 68.64 \\
         MADD \cite{zhao2021multi} & 65.65 & 68.64 \\
         RECCE \cite{cao2022end} & 64.31 & 68.71 \\
         PEL \cite{gu2022exploiting} & 67.39 & 69.18 \\
         MSWT \cite{liu2022multi} & 68.71 & - \\
         MSFnet \cite{guo2023data} & - & 69.50 \\
         SFGD \cite{wang2023dynamic} & \underline{69.27} & \underline{75.83} \\
         \textbf{Ours} &\textbf{71.60}	&\textbf{75.99} \\
    \bottomrule
  \end{tabular}
    \caption{Cross-dataset comparison in terms of AUC ($\%$) training on the FF++ (LQ) dataset. Best results are marked in bold, and the second best are marked with underline.}
    \label{tab:cross-1}
\end{table}

\begin{table}
  \centering
  \belowrulesep=0pt
  \aboverulesep=0pt
  \begin{tabular}{c|c|c}
    \toprule
    \multicolumn{1}{c|}{Training DataSet} & \multicolumn{2}{c}{WildDeepfake} \\
    \midrule
        Method & FF++ (LQ) & Celeb-DF (v2) \\
         \midrule
         Xception \cite{chollet2017xception} &	59.20 &	77.91 \\
         Add-Net \cite{zi2020wilddeepfake} & 53.88 & 62.12 \\
         CNNDetection \cite{wang2020cnn} & 59.66 & 76.27 \\
         GANFengerprient \cite{frank2020leveraging} & 53.98 & 67.08 \\
         F3Net \cite{qian2020thinking} & 55.95 & 60.88 \\
         Multi-att \cite{zhao2021multi} & - & 76.95 \\
         RFM \cite{wang2021representative} & 57.63 & 62.38 \\
         HFF \cite{luo2021generalizing} & \underline{62.76} & 76.95 \\
         MSVT \cite{lin2022improved} & - & 74.91 \\
         RECCE \cite{cao2022end} & 59.86 & - \\
         PEL \cite{gu2022exploiting} &	61.60 &	\underline{82.94} \\
         \textbf{Ours} &\textbf{71.94}	&\textbf{91.20} \\
    \bottomrule
  \end{tabular}
    \caption{Cross-dataset comparison in terms of AUC ($\%$) training on the WildDeepfake dataset. Best results are marked in bold, and the second best are marked with underline.}
    \label{tab:cross-2}
\end{table}

\subsection{Experimental Setup}
\textbf{Datasets.} We conducted extensive experiments on three public datasets: FaceForensics++ (FF++)~\cite{rossler2019faceforensics++}, Celeb-DF (v2)~\cite{li2020celeb}, and WildDeepfake~\cite{zi2020wilddeepfake}. FaceForensics++ is a well-known dataset for face forgery detection, which contains 1,000 real videos and 4,000 forged videos generated by four forgery methods: DeepFakes (DF), Face2Face (F2F), FaceSwap (FS), and NeuralTextures (NT). Celeb-DF (v2) consists of 590 real videos collected from YouTube and 5,639 corresponding fake videos. Compared to FaceForensics++, videos in Celeb-DF (v2) have better forgery quality and fewer visual artifacts. WildDeepfake is a challenging real-world face forgery dataset. It contains 3,805 real face sequences and 3,509 forged face sequences collected from the Internet. Since these video sequences are real-world data collected from the Internet through different sources, the WildDeepfake dataset is more diverse in forgery methods, video duration, video quality, and video content.


\noindent
\textbf{Evaluation Metrics.} For a fair comparison with existing state-of-the art methods, we employ Accuracy (Acc) and Area Under the Receiver Operating Characteristic Curve (AUC) as our evaluation metrics.

\noindent
\textbf{Implementation Details.} We use Dlib\footnote{Dilb https://pypi.org/project/dlib/} for face detection and alignment and unify the size of all images to $ 224\times224 $. The backbone network is trained using the AdamW~\cite{loshchilov2018fixing} optimizer with an initial learning rate set to $ 1.2\times10^{-3} $ and the weight decay set to $ 0.05 $.

\begin{table*}
\begin{center}
\belowrulesep=0pt
\aboverulesep=0pt
\renewcommand{\arraystretch}{1.2}
\begin{tabular}{c|c|c|cccccc}
\toprule
\multirow{2}{*}{Method} &
\multirow{2}{*}{SSAA} &
\multirow{2}{*}{FQMP} &

\multicolumn{2}{c}{FF++ (HQ)} &

\multicolumn{2}{c}{WildDeepfake} &

\multicolumn{2}{c}{Celeb-DF (v2)} \\
\cline{4-5}
\cline{6-7}
\cline{8-9}
  & & & Acc ($\%$) & AUC ($\%$)  & Acc ($\%$) & AUC ($\%$)   & Acc ($\%$) & AUC ($\%$) \\
\midrule
a&×&× &	97.78&	99.14&	86.35&	92.46&	99.22&	99.99 \\
b& \Checkmark &× &	96.67&	98.94&	87.22&	93.15&	99.41&	100 \\
c& × & \Checkmark & 98.52&	99.92&	86.97&	91.09&	100&	100  \\
\textbf{Ours}& \Checkmark & \Checkmark &	98.89&	99.67&	88.34&	92.33&	100&	100  \\
\bottomrule
\end{tabular}
\end{center}
\caption{Ablation study on the WildDeepfake dataset. (a) The backbone network w/o quality mapping and SSAA. (b) The proposed network w/o quality mapping. (c) The proposed framework w/o SSAA.}
\label{tab:ablation}
\end{table*}

\subsection{Experimental Results}
\noindent
\textbf{Intra-dataset Evaluation.} We compare our method with state-of-the-art approaches on three well-known datasets: FaceForensics++, Celeb-DF (v2), and WildDeepfake. \Cref{tab:intro} shows that our method outperforms other reference methods on almost all datasets. Specifically, our method achieves an accuracy of 98.89$\%$ on the FF++ (HQ) dataset, outperforming RECCE~\cite{cao2022end} by 1.83$\%$. RECCE trains a reconstruction network using real face images and utilizes the hidden layer features of the reconstruction network for classification. Unlike it, our approach aims to explore the distinctions among face images of different qualities and differentiate between real and fake face data at a finer granularity.
SFDG~\cite{wang2023dynamic} introduces a novel spatial frequency dynamic graph network that effectively captures subtle forgery clues by exploiting the relationship between spatial and frequency domains. Through carefully designed modules, including content-guided adaptive frequency extraction and dynamic graph-based feature fusion, this method has achieved good results on various benchmarks. In comparison, our method improves the accuracy of the more challenging real-word dataset WildDeepfake by 3.93$\%$, which significantly outperforms this method and other compared methods. In addition, we evaluate our method on the Celeb-DF (v2) dataset, which contains videos with better visual quality. Our proposed method also achieves the best results on the Celeb-DF (v2) dataset in terms of Acc and AUC, which shows that our method still has excellent performance on this high-quality deepfake dataset. These experimental results demonstrate the effectiveness and superiority of our DeepFidelity method.

~\\
\noindent
\textbf{Cross-dataset Evaluation.} To evaluate the generalization ability of our proposed method, we conduct cross-dataset experiments by testing on unseen datasets. Specifically, we first train the model on the FF++ (LQ) dataset and test it on the WildDeepfake and Celeb-DF (v2) dataset. The experimental AUC results are shown in \cref{tab:cross-1}. Qian \etal~\cite{qian2020thinking} addresses the challenge of detecting face forgeries in compressed images and videos using frequency-aware techniques. The proposed F3-Net framework leverages frequency-based insights, including decomposed image components and local frequency statistics, through a two-stream collaborative learning approach. This method achieves good results in detecting low-quality forgeries in the FaceForensics++ dataset, but the generalization performance on unknown datasets could be better.
As can be seen from the table, it is evident that our proposed DeepFidelity method significantly outperforms all listed methods on previously unseen test datasets, achieving AUC of 71.60$\%$ on the WildDeepfake dataset and 75.99$\%$ on the Celeb-DF (v2) dataset.

Meanwhile, we also conduct the cross-dataset evaluation by training on the WildDeepfake dataset and testing on the FF++ (LQ) and the Celeb-DF (v2) datasets. The experimental results are shown in \cref{tab:cross-2}. It can be observed that the results of our proposed method on the unknown datasets outperform the other listed methods. 
\cite{gu2022exploiting} introduces a Progressive Enhancement Learning (PEL) framework for detecting facial forgery. It leverages fine-grained frequency information and combines RGB decomposition with self-enhancement and mutual-enhancement modules to effectively capture and enhance forgery traces. PEL is comparable to our method on the WildDeepfake dataset but exhibits poor generalization capabilities. 
In comparison, our method outperformed PEL by 10.34$\%$ on the unknown dataset FF++ (LQ) and 8.26$\%$ on Celeb-DF (v2). The results in \cref{tab:cross-1,tab:cross-2} demonstrate that perceptual forgery fidelity is a robust forgery clue that enables good generalization ability.

\begin{figure*}
  \centering
    \includegraphics[width=1.0\linewidth]{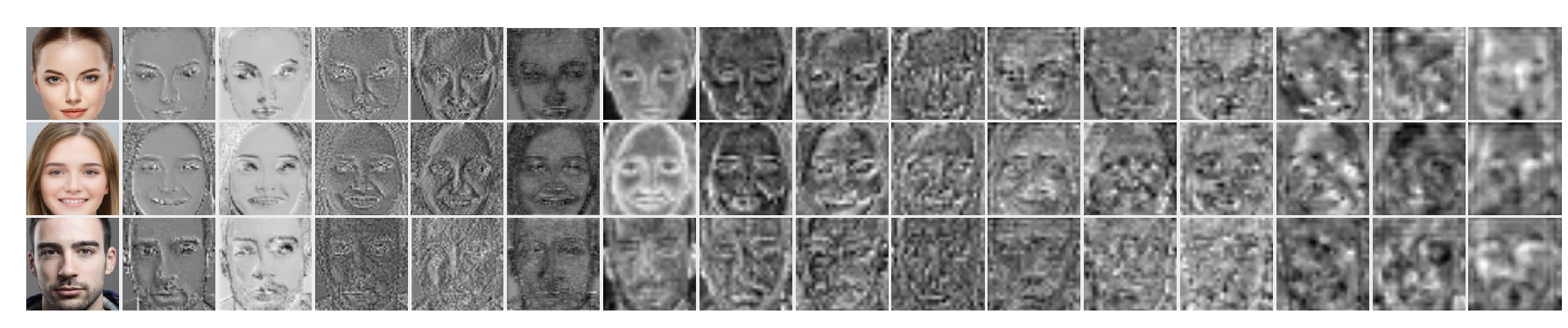}
  \caption{Visualization of face feature maps. We visualize the output feature maps of the first 15 blocks of the network, from which we can see that as the number of network layers deepens (from left to right), the semantic features extracted by the network gradually become abstract. In order to better exploit the symmetric spatial relationship in the feature maps, we only apply the SSAA in the shallow layer.}
  \label{fig:featuremap}
\end{figure*}

\subsection{Ablation Study}

\textbf{Effectiveness of Proposed Quality Mapping and SSAA.} By assessing face quality scores, our method maps face data of different qualities into distinct score ranges. Additionally, the Symmetric Spatial Attention Augmentation based vision Transformer is designed to model the geographic long-distance relationship for augmenting local features. To verify the effectiveness of different components in our framework, we conduct the following experiments on the WildDeepfake dataset: (a) The baseline model w/o quality mapping and SSAA. (b) The proposed framework w/o quality mapping. (c) The proposed framework w/o SSAA. 
The experimental results are shown in \cref{tab:ablation}.

By comparing the results of (a) and (b), it can be observed that the SSAA we proposed enhances the model performance on most datasets, especially on the WildDeepfake dataset, where the Accuracy increased by 0.87$\%$ and the AUC improved by 0.69$\%$. This suggests that leveraging the symmetry of face data for geographic long-distance relationships modeling and local feature augmentation contributes to the model learning more expressive feature representations. Contrasting (a) and (c), 
it is evident that mapping face data of different qualities into distinct scores leads to significant gains in both Acc and AUC due to finer data granularity. When all components are combined, our proposed approach achieves the best performance in terms of Acc across all datasets. The accuracy metric improved by 1.11$\%$ on the FF++ (HQ) dataset, by 1.99$\%$ on the WildDeepfake dataset, and by 0.78$\%$ on the Celeb-DF (v2) dataset compared to the baseline.

\begin{figure}
  \centering
  \begin{subfigure}{0.48\linewidth}
    \includegraphics[width=1.0\linewidth]{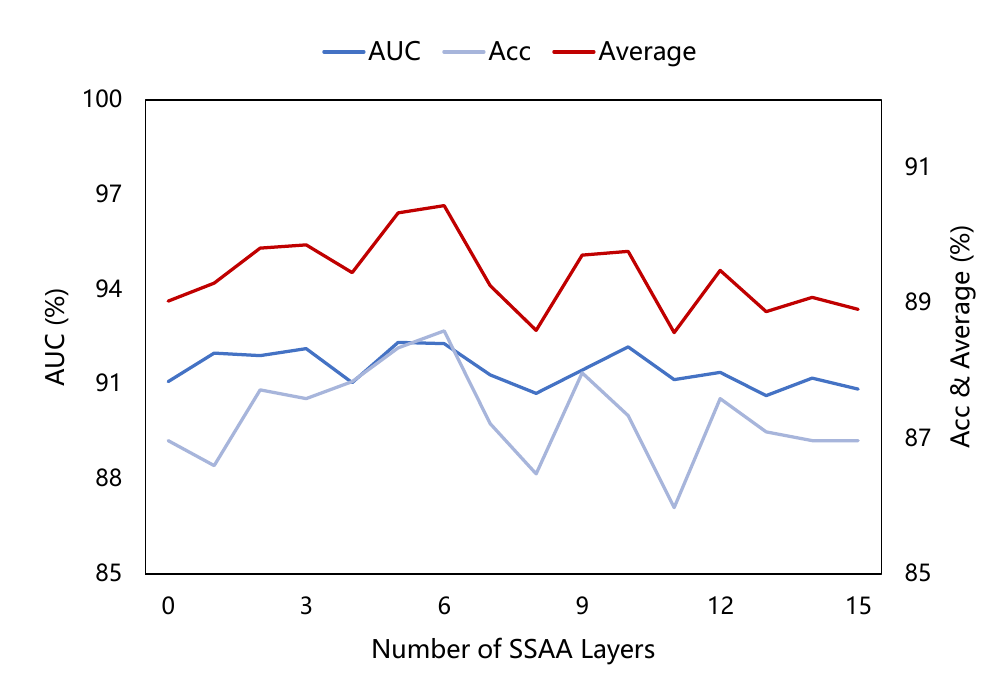}
    \caption{Experimental results on the WildDeepfake Dataset.}
    \label{fig:layer-selection-a}
  \end{subfigure}
  \hfill
  \begin{subfigure}{0.48\linewidth}
    \includegraphics[width=1.0\linewidth]{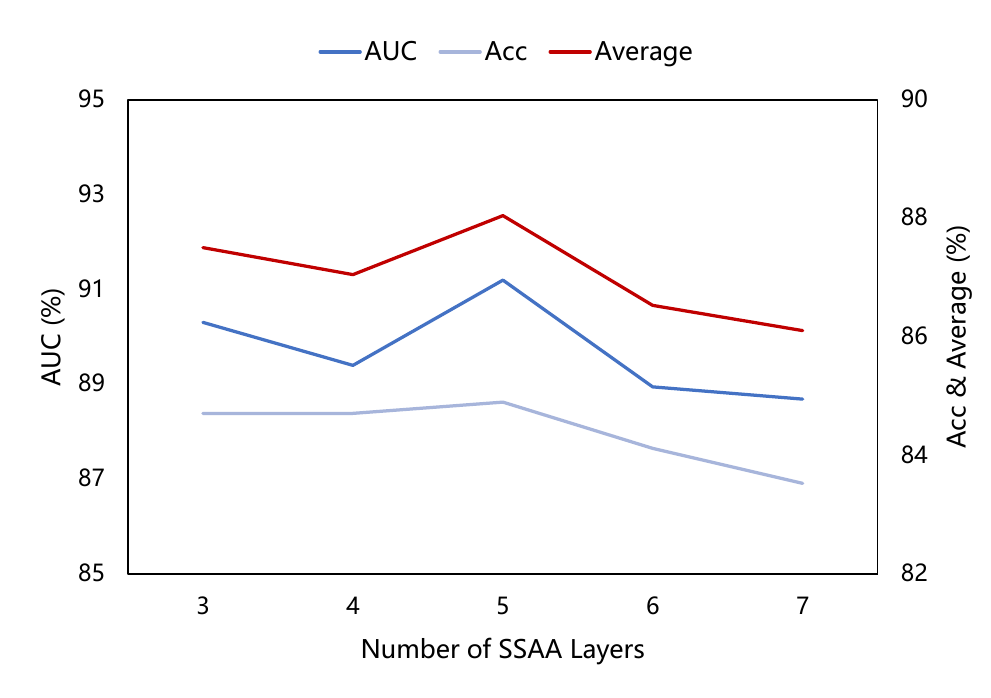}
    \caption{Cross-domain testing on the Celeb-DF (v2) dataset.}
    \label{fig:layer-selection-b}
  \end{subfigure}
  
  \caption{Experimental results of applying SSAA on different number of layers in terms of Acc ($\%$), AUC ($\%$) and average of them.}
  \label{fig:layer-selection}
\end{figure}

\noindent
\textbf{Setting of SSAA Layers.} To further evaluate the impact of varying the number of SSAA layers on the experimental results, we conduct a comprehensive series of experiments on the WildDeepfake dataset. The results, as shown in \cref{fig:layer-selection-a}, distinctly indicate that employing an insufficient number of SSAA layers cannot adequately model geographic long-distance relationships. Furthermore, a detailed examination of the feature maps generated by the first 15 blocks of the network revealed a noteworthy trend. As depicted in \cref{fig:featuremap}, as the depth of the network increases, the receptive field undergoes a progressive enlargement, while positional information undergoes a corresponding reduction. This transition process leads to features extracted by the network becoming more abstract. Consequently, in the deeper strata of the network, the once prominent local symmetric relationships gradually diminish. To further explore the appropriate number of SSAA layers, we conduct cross-domain testing on the Celeb-DF (v2) dataset. The results in \cref{fig:layer-selection-b} show that the best performance is achieved when the first five layers are selected for symmetric feature augmentation. Taking these observations into consideration, we decide to employ symmetric spatial attention augmentation in the first stage of the model, i.e., the first five blocks.

\noindent
\textbf{Quality Assessment Methods.} To explore the impact of different quality assessment methods on our approach, we also conduct experiments using two additional no-reference image quality assessment methods, BRISQUE and NIQE, as our quality scoring models. The experimental results on the WildDeepfake dataset are presented in \cref{tab:iqa-model}. It can be observed that the accuracy achieved when using Ser-fiq as the face quality assessment model is somewhat superior to the other two methods. BRISQUE and NIQE are designed to evaluate the quality of natural images, not specifically tailored for face data. Ser-fiq aims to predict the suitability of face images for face recognition systems, making it more suitable for our data and task. The results above indicate that using a better quality assessment model contributes to improving the performance of our method.

\noindent
\textbf{Analysis on Face Images with Different Qualities.} To explore the detection performance of our method on face data of different qualities, we partitioned the real and fake face images in the WildDeepfake dataset into four groups based on their face quality scores. The experimental results of our method and binary classification without quality score mapping are presented in \cref{tab:quality-range}. It can be observed that without quality score mapping, the model tends to misclassify real images as fake, leading to low accuracy on real images. In contrast, our method achieves a more equitable classification, with accuracy distributed more uniformly across different quality intervals.

\begin{table}
  \centering
  \begin{tabular}{ccc}
    \toprule
        IQA Model & Acc & AUC \\
         \midrule
         Ser-fiq \cite{terhorst2020ser} & 88.34 & 92.33 \\
         BRISQUE \cite{mittal2012no} & 87.34 & 91.90 \\
         NIQE \cite{mittal2012making} & 87.59 & 92.91 \\
    \bottomrule
  \end{tabular}
    \caption{Experimental results in terms of Acc ($\%$) and AUC ($\%$) for using different image quality assessment methods in face quality mapping phase.}
    \label{tab:iqa-model}
\end{table}

\begin{table}
  \centering
  \belowrulesep=0pt
  \aboverulesep=0pt
  \begin{tabular}{c|c|cc}
    \toprule
        Class & Quality Range & Binary classification & Ours \\
         \midrule
         \multirow{4}{*}{Fake} & 0.00 - 0.25 & 80.39 & 80.11 \\
         & 0.25 - 0.50 & 85.86 & 83.00 \\
         & 0.50 - 0.75 & 88.32 & 83.48 \\
         & 0.75 - 1.00 & 93.31 & 86.89 \\
         \midrule
         \multirow{4}{*}{Real} & 0.00 - 0.25 & 89.05 & 86.32 \\
         & 0.25 - 0.50 & 77.02 & 81.69 \\
         & 0.50 - 0.75 & 79.63 & 88.76 \\
         & 0.75 - 1.00 & 72.96 & 86.17 \\
        \bottomrule
  \end{tabular}
    \caption{Experimental results in terms of Acc ($\%$) for face images in different quality ranges.}
    \label{tab:quality-range}
\end{table}

%% file: sec/5_conclusion.tex
\section{Conclusion}
\label{sec:conclusion}
In this paper, we introduce a novel approach for Deepfake detection named DeepFidelity. Considering that face data of different qualities may bring different levels of detection difficulty to the model, we propose the concept of perceptual forgery fidelity. This enhances the model's capability to learn complex samples by mapping real and fake face data of different qualities to distinct scores. In addition, we design a network architecture called SSAAFormer, which models the geographic long-distance symmetry relationship of faces in the shallow layers of the network to augment local features, enabling the model to learn more expressive feature representations. Intra-dataset and cross-dataset evaluation results demonstrate the superiority and robustness of the proposed DeepFidelity compared to state-of-the-art methods. In the future, we will explore face quality assessment methods that are more suitable for our approach and evaluate the proposed method in more realistic and unconditional forgery detection situations. Additionally, we will work on more advanced fidelity regression models to achieve better performance.